\documentclass[11pt]{article}
\RequirePackage{xcolor,graphicx}
\usepackage{coling2020}
\usepackage{algorithm, algorithmicx}
\usepackage{algpseudocode}%
\usepackage{times}
\usepackage{hyperref}

\usepackage{wrapfig}

\usepackage{latexsym,amsmath,amsfonts,amssymb,bm}
\usepackage[normalem]{ulem}

\makeatletter
\g@addto@macro{\normalsize}{%
\setlength{\abovedisplayskip}{0pt}%
\setlength{\abovedisplayshortskip}{0pt}%
\setlength{\belowdisplayskip}{0pt}%
\setlength{\belowdisplayshortskip}{0pt}}
\makeatother

\def\complex{CX}
\def\sys{\textsc{Complex-V2}}

\def\sysS{\textsc{SimplE-V2}}

\def\R{\mathbb{R}}

\def\bs{\bm{e_s}}
\def\br{\bm{w_r}}
\def\bo{\bm{e_o}}
\def\bhs{\bm{h_s}}

\def\br{\bm{r}}
\def\bto{\bm{t_o}}

\newcommand{\ee}{e}
\newcommand{\es}{s}
\newcommand{\eo}{o}
\newcommand{\er}{r}

\def\R{\mathbb{R}}

\def\bs{\bm{s}}
\def\br{\bm{r}}
\def\bo{\bm{o}}

\DeclareMathOperator{\phitranse}{\ensuremath{\phi_\text{TransE}}}
\DeclareMathOperator{\phirotate}{\ensuremath{\phi_\text{RotatE}}}
\DeclareMathOperator{\phidistmult}{\ensuremath{\phi_\text{DM}}}
\DeclareMathOperator{\phicomplex}{\ensuremath{\phi_\text{CX}}}

\newcommand\norm[1]{\left\lVert#1\right\rVert}

\colingfinalcopy 

\title{Knowledge Base Completion: Baseline Strikes Back (Again)}

\author{
Prachi Jain$^{1}$, 
{Sushant Rathi}$^1$, 
Mausam$^1$ {\normalfont and} 
Soumen Chakrabarti$^2$
\\ 
$^1$ Indian Institute of Technology Delhi \\
$^2$ Indian Institute of Technology Bombay  \\
\{p6.jain, rathisushant5\}@gmail.com,
mausam@cse.iitd.ac.in,
soumen.chakrabarti@gmail.com
}

\date{}

\begin{document}
\maketitle
\begin{abstract}

KBC datasets have a large number of (positive) training instances and an even larger number of negative training instances via negative sampling.
Various KBC methods define diverse loss functions. 
These have different computational costs and memory footprints.
Given the resource constraints of a computational environment (e.g. GPU size and practical training time available), 
these resource requirements dictate the extent of negative sampling that is feasible.
Sometimes, apparently minor changes in the design of the loss function can change system accuracy to a surprising extent, and also open up paths to computational optimizations. To our knowledge, these trade-offs have not been studied adequately.
One consequence is that baselines have remained unnecessarily weak
\cite{baseline-KadlecBK17,trick-iclr20-Ruffinelli}.
In this paper, we find that careful attention to these details can considerably improve old baselines, sometimes surpassing models and methods proposed since.
Specifically, if large numbers of negative triples can be accommodated, the basic ComplEx model is extremely competitive. We call this approach \sys. We also highlight how various KBC methods, recently proposed in the literature, benefit from this training regime and become indistinguishable in terms of performance on most datasets. Our work calls for a reassessment of their individual value, in light of these findings.

\end{abstract}

\section{Introduction}
\label{sec:Intro}

A Knowledge base (KB) is a collection of world knowledge facts in the form of a triple where the subject ($s$) is related to object ($o$) via relation ($r$), e.g.: $\langle$Donald Trump, is President of, USA$\rangle$. Most KBs are incomplete --- the Knowledge Base Completion (KBC) task infers missing facts from the known triples, hence making them more effective for end tasks like search and question answering.

Various translation \cite{BordesUGWY2013TransE,RotatESunDNT19,JiHXLZ2015TransD,RotatESunDNT19}, multiplicative \cite{yang15export:241703,TrouillonWRGB2016Complex,Lacroix2018canonical,JainKMC2018TypeDmCx,balazevic2019tucker,kazemi2018simple}, and deep learning based \cite{KBGAT2019,dettmers2017convolutional,ConvKBNguyenNNP18,NN-SchlichtkrullKB18}, approaches for KBC have been discussed in the literature. The scoring functions of these methods takes in the embeddings of $s$, $r$, and $o$ to generates a score indicating the confidence in the truthfulness of the fact. 

\newcite{dettmers2017convolutional} showed that 1-N scoring i.e. computing all possible fact variations -- $(s,r,*)$ and $(*,r,o)$, at the same time, can improve model performance, with faster convergence as well. We leverage this idea to train older models with large (perhaps all entities) negative sample set to match those of recent state-of-the-art (SOTA) models. We find that this training method is difficult to incorporate when we switch from $L_1$ to $L_2$ norm; in particular, translation models such as TransE, RotatE, where 1-N scoring does not scale. 
We resort to the trick of gradient accumulation to train translation models with all entities as negative samples, at the cost of increased training time. A majority of recently released models, as well as older models such as SimplE and Complex, can improve their performance significantly when trained with large (exhaustive) negative samples 
(we refer to the improved models as \sysS{} and \sys{} respectively). To our surprise, \sys{} outperforms or is very close in performance to all recent state of the art KBC models!

In light of these findings, we draw two conclusions:  (1)~there is a need to reassess the value offered by recent KBC models against the older Complex (also called \complex{} here) model, and (2) any new KBC model must compare against the baseline of \sys{} to demonstrate empirical gains. 

Moreover, as long as scalable, all KBC models must use the training regime of 1-N scoring to be able to use all entities as negative samples and obtain a superior performance. For models where 1-N scoring does not scale (such as RotatE), gradient accumulation over multiple batches can be used at the cost of increased training time.

We will release an open-source implementation 
of the models 
for further exploration.

\section{Background and Related Work}
\begin{wraptable}{r}{0.5\linewidth}
\centering
\begin{small}
\begin{tabular}{|l|l|}
\hline
\textbf{Model ($M$)} & \textbf{Scoring function ($\phi^M$)} \\ \hline
TransE~\protect\shortcite{BordesUGWY2013TransE} & $-\norm{\bs + \br - \bo}_2$ \\ \hline
RotatE~\protect\shortcite{RotatESunDNT19} & $-\norm{\bs \bullet \br - \bo}_2$ \\ \hline \hline

ComplEx~\protect\shortcite{TrouillonWRGB2016Complex} & $\Re\langle \bs, \br, \bo^\star\rangle$ \\ \hline
SimplE~\protect\shortcite{kazemi2018simple} & $ \frac{1}{2}(\langle \bhs, \br, \bto \rangle + \langle \bto, \br^{-1}, \bhs \rangle)$ \\ \hline


\end{tabular}
\end{small}
\caption{Scoring functions for KBC models. First 1-2 rows list translation models, row 3-5 lists bilinear models, row 6 list a deep learning models. Larger value implies more confidence in the validity of the triple. $\bullet$ denotes rotation. $\star$ denotes the complex conjugate. $\Re$ refers to the real part of the complex valued score returned by the ComplEx model.} \label{model_scoring_function}

\end{wraptable}

We are given an incomplete KB with entities $\mathcal{E}$ and relations $\mathcal{R}$.  The KB also contains $\mathcal{T}=\{\langle s, r, o \rangle\}$, a set of known valid tuples, each with subject and object entities $s,o \in \mathcal{E}$, and relation $r \in \mathcal{R}$. The goal of KBC model is to predict the validity of any tuple not present in~$\mathcal{T}$. Previous approaches fit continuous representations (loosely, ``embeddings'') to entities and relation, so that the belief in the veracity of $\langle s, r, o \rangle$ can be estimated as an algebraic expression (called a scoring function $\phi$) involving those embeddings. The scoring functions for the models considered in this work are outlined in Table-\ref{model_scoring_function}. The embedding of $s$, $o$ and $r$ are denoted as $\bs, \bo, \br$ respectively. 


\noindent
\subsection{TransE}

TransE \cite{BordesUGWY2013TransE} embeds each entity $\ee$ (variously, subject $\es$ or object $\eo$) to vectors $\bs, \bo \in \mathbb{R}^D$ and relations $\er$ to vectors~$\br\in \mathbb{R}^D$ as well.  The score function is defined as 
\begin{align}
    \phitranse(\es,\er,\eo) &= \| \bs + \br - \bo \|_1.
\end{align}
As we shall see, the use of $L_1$ norm above is significant; if it is replaced by the $L_2$ norm, there are significant implications for memory footprint.  To design the loss function, we observe that if $(\es,\er,\eo)\in\text{KB}$, we want $\phitranse(\es,\er,\eo)$ to be small; otherwise, if $(\es,\er,\eo)\not\in\text{KB}$, we want it to be large.  These considerations are combined into the loss function using a margin $\Delta>0$ for the negative triples:
\begin{align}
    \mathcal{L}_\text{TransE} &=
    \sum_{(\es,\er,\eo)\in\text{KB}} \phitranse(\es,\er,\eo)
    + \sum_{(\es',\er',\eo')\not\in\text{KB}} \big[ \Delta - \phitranse(\es',\er',\eo') \big]_+,
\end{align}
where $[a]=\max\{0,a\}$ is the ReLU or hinge function.  For each $(\es,\er,\eo)\in\text{KB}$, the number of perturbations $(s',r',o')$ that are (assumed to be) not in the KB is astronomically large, possibly approaching $E^2R$, where the KB has $E$ distinct entities and $\er$ distinct relation types.  Usually, negative sampling is limited to perturbing only $\es$ or only $\eo$, resulting in $O(E)$ negative triples per positive triple, at most.  But even this is considered computationally too burdensome, and a random sample is drawn.

\subsection{RotatE}

Computationally, RotatE \cite{RotatESunDNT19} is similar to TransE.
It places $\bs, \br, \bo \in \mathbb{C}^D$ in the complex space, enforces unit complex modulus $|\br^d|=1$ for each element of $\br$, and defines 
\begin{align}
\phirotate &= - \sum_{d\in[D]} |\bs^d \br^d - \bo^d|.
\end{align}
Here $|c|$ is the modulus of complex number~$c\in\mathbb{C}$ and $\bs^d\br^d$ is the product of two complex numbers.  Note that the sum is similar to an $L_1$ distance.  RotatE can learn symmetry vs.\ antisymmetry, inversion and composition, and generally performs better than TransE.  Because of the difference to be computed for each dimension $d$, we will regard TransE and RotatE as members of the \emph{additive} family of KBC methods.

\subsection{DistMult and ComplEx}

In contrast to additive KBC methods, DistMult and ComplEx \cite{yang15export:241703,TrouillonWRGB2016Complex,Lacroix2018canonical,JainKMC2018TypeDmCx,balazevic2019tucker,kazemi2018simple} can be regarded as \emph{multiplicative} methods, for reasons clarified below.  For DistMult, $\bs, \br, \bo \in \mathbb{R}^D$ and 
\begin{align}
\phidistmult(\es,\er,\eo) &= \sum_{d\in [D]}  \bs^d \br^d \bo^d.
\end{align}
For ComplEx, $\bs, \br, \bo \in \mathbb{C}^D$ and
\begin{align}
\phicomplex(\es,\er,\eo) &= \R\left[
\sum_{d\in [D]}  \bs^d \br^d \bo^{d\star}\right],  \label{eq:complexv2}
\end{align}
where $c^\star$ is the conjugate of complex number~$c$ and $\R(c)$ is its real part.  In both cases, observe that there is no addition or subtraction inside the sum over dimensions~$d$.  This has important implications.

For either DistMult or ComplEx, the loss is commonly defined as
\begin{align}
\mathcal{L} &= - \!\! \sum_{( \es,\er,\eo )\in\mathcal{T}_{tr}} \!\! \Bigl( \log\Pr(\eo|\es, \er; \theta) + \log\Pr(\es|\eo,\er; \theta) \Bigr)
\label{baseline:eq:type_SoftMaxLoss}
\intertext{where}
\Pr(\eo|\es,\er) &= \frac{\exp( \phi(\es,\er,\eo))}{\sum_{o'} \exp( \phi(\es,\er,\eo'))}
\quad\text{and}\quad
\Pr(\es|\eo,\er) = \frac{\exp( \phi(\es,\er,\eo))}{\sum_{s'} \exp( \phi(\es',\er,\eo))},
\label{baseline:prob}
\end{align}
Here, again, observe the potential performance bottleneck of the sums in the denominator ranging over $E$ entities.  Indeed, early implementations approximated the full sum in the denominator with a partial sum over a random subset of terms, suitably scaled, plus the numerator itself (to maintain consistency).  As we shall see, this sampling approximation may not be necessary; the specific inner-product form \eqref{eq:complexv2} lets us evaluate the full denominator sum efficiently.


\section{Additive vs.\ multiplicative loss with all negative triples} \label{baseline:sec:addvsmul}

Here we first describe how \eqref{baseline:prob} can be fully evaluated without sampling approximation, supported by highly efficient tensorized computation libraries.  Then we describe why this appears more difficult for additive formulations (TransE and RotatE).  Finally, we describe how tweaking the distance norm from $L_1$ to $L_2$ might open up TransE to  the same efficiency.  In experiments, however, we see that $L_1$ norm generally gives more accurate KBC models.

\subsection{Inner product}

\newcite{dettmers2017convolutional} suggested taking one $(s,r)$ pair and scoring it against all $E$ entities $\eo$ in a batch method they called ``1-N scoring'', instead of computing the score of one fact $( \es,\er,\eo )$ at a time, that they called ``1-1 scoring''.  For any {\it multiplicative} method in general, the score for all entities can be computed in parallel via a simple matrix multiplication, which is both memory and time-efficient, thanks to the optimized implementations provided in BLAS libraries.

To get into more detail, let us ask how the computation in \eqref{baseline:prob} can be efficiently vectorized in case of DistMult.  $\phidistmult$ is often written in the form $(\bs\odot \br) \cdot \bo$, where $\odot$ is elementwise product, and $\cdot$ is an inner product.  Here $(\bs\odot\br)\in \mathbb{R}^D$, as is $\bo\in \mathbb{R}^D$.  If we want to evaluate $\phidistmult(\es,\er,\eo)$ over all $o \in \mathcal{E}$, we can write it  as a matrix-vector product between a $E\times D$ entity embedding matrix $\mathbf{E}$ and a $D\times1$ row vector $\bs\odot\br$ followed by a sum aggregation:
\begin{align}
    \sum_{o\in [E]} \exp\Big( \mathbf{E}_{E\times D} \; (\bs\odot\br)_{D\times 1}
    \Big)_{E\times 1}.
\end{align}
(In reality we would use log-sum-exp for numerical stability, but the computation structure will remain the same.)  The total intermediate space needed to compute the above expression is $O(DE)$.  If we want to further batch up subject entities $\es$ in batches of size $B$, the space required is $O(BD + DE)$.

\subsection{$L_1$ distance}

Let us now shift focus to $\phitranse$.  As with $\bs\odot\br$ in DistMult, pre-computation of $\bs+\br$ creates no trouble, giving a $D$-dimensional vector.  If we have a batch of $B$ $(s,r)$ pairs, this gives us a $B\times D$ matrix; call this $\mathbf{B}$.  The other matrix of interest is $\mathbf{E} \in \mathbb{R}^{E\times D}$ as before.  Effectively, we have a set of $B$ points in $D$ dimensions, another set of $E$ points in $D$ dimensions, and we wish to compile a $B\times E$ matrix $\mathbf{A}$ of pairwise $L_1$ distances.

In non-vectorized code, this is easily possible to compute within $O(BD+DE)$ input space, $O(BE)$ output space, and no (or $O(1)$) working space, using the following approach.

\begin{algorithmic}[1]
\For{$b \in [B]$} 
\For{$e \in [E]$}
\State $\mathbf{A}[b,e] \leftarrow 0$
\For{$d \in [D]$}
\State $\mathbf{A}[b,e] \leftarrow \mathbf{A}[b,e] +
\underbrace{\Big|\mathbf{B}[b,d] - \mathbf{E}[e,d]\Big|}_{}$
\EndFor
\EndFor
\EndFor
\end{algorithmic}

Structurally, this is identical to the non-vectorized code for matrix multiplication.

\begin{algorithmic}[1]
\For{$b \in [B]$} 
\For{$e \in [E]$}
\State $\mathbf{A}[b,e] \leftarrow 0$
\For{$d \in [D]$}
\State $\mathbf{A}[b,e] \leftarrow \mathbf{A}[b,e] +
\underbrace{\mathbf{B}[b,d] \; \mathbf{E}[e,d]}_{}$
\EndFor
\EndFor
\EndFor
\end{algorithmic}

SciPy defines a general library function \texttt{scipy.spatial.distance.cdist}\footnote{\protect\url{https://docs.scipy.org/doc/scipy/reference/generated/scipy.spatial.distance.cdist.html}}, which allows the compilation of $B\times E$ distances using any $L_p$ norm.  $L_1$ corresponds to input parameter \texttt{metric='cityblock'}, but the default is $L_2$, or \texttt{metric='euclidean'}.

To our astonishment, despite the similarity in the above pseudocodes, the memory complexity of a vectorized version of $L_1$ distance is $O(BDE)$, instead of $O(BD+DE+BE)=O(BD+DE)$ achievable for matrix multiplication.  It appears the structure of computation changes to the following.

\begin{algorithmic}[1]
\State allocate a $B\times E \times D$ tensor $\mathbf{X}$
\For{$d \in [D]$}
\State $\mathbf{X}[:,:, d] \leftarrow \mathbf{B}[:,d] \otimes \mathbf{E}[:,d]
\in \mathbb{R}^{B\times E \times 1}$
\Comment{Each layer $d$ gets a $B\times E$ outer product.}
\EndFor
\State $\mathbf{A} \leftarrow \operatorname{sum-reduce}_d(\mathbf{X}) \in \mathbb{R}^{B\times E}$  \Comment{Aggregate across layers $d$.}
\end{algorithmic}

Needless to say, this wastes a lot of space, making it difficult to deal with all negative subject or object entities, even for small batches.

\subsection{The special case of $L_2$ distance}

SciPy's default choice of $L_2$ distances in their \texttt{cdist} routine affords a space-efficient solution.  Suppose we have a matrix $\bm{B} \in \R^{B\times D}$ and a matrix $\bm{Y} \in \R^{E\times D}$. If we want a $B\times E$ matrix $\bm{A}$ with dot products, i.e., $\bm{A}[b,e] = B[b,:] \cdot Y[e,:]$, we can efficiently compute $\bm{A} = \bm{B} \bm{Y}^\top$. What if we want $\bm{A}[b,e] = \| \bm{B}[b,:] - \bm{E}[e,:] \|_2^2$? 
We can write this as 
\begin{align*}
\bm{A}[b,e] = \|\bm{B}[b,:] - \bm{E}[e,:] \|_2^2 
&= (\bm{B}[b,:] - \bm{E}[e,:]) \cdot (\bm{B}[b,:] - \bm{E}[e,:]) \\
&= \|\bm{B}[b,:]\|^2_2 + \|\bm{E}[e,:]\|^2_2 - 2 \bm{B}[b,:]\cdot \bm{E}[e,:]
\end{align*}
We can compute the first two terms without any asymptotic increase in storage, and the third term is the same as in case of dot product.  If we need
\begin{align*}
\bm{A}[b,e] &= \| \bm{B}[b,:] - \bm{E}[e,:] \|_2  \qquad \text{(i.e., $L_2$, not $L_2$-squared)}
\intertext{we can write this as}
\bm{A}[b,e] &= \sqrt{\|\bm{B}[b,:]\|^2_2 + \|\bm{E}[e,:]\|^2_2 - 2 \bm{B}[b,:]\cdot \bm{E}[e,:]}.
\end{align*}
I.e., there is a final elementwise square-root on the $B\times E$ matrix.  The gradient changes, but not the basic space and time performance structure.  

This trick can be used if we change the formulation of TransE and RotatE to use $L_2$ distances instead of $L_1$ distances:
\begin{align}
\phitranse(\es,\er,\eo) &= -\norm{\bs + \br - \bo}_2 \quad  \text{and} \\
\phirotate(\es,\er,\eo) &= -\norm{\bs \odot \br - \bo}_2.
\end{align}
Without experimental evaluation, we do not  know the impact of the change of norm on the predictive accuracy of these models.  We end this chapter with such an evaluation, which shows that switching from $L_1$ to $L_2$ is unfortunately detrimental to predictive accuracy. However, ComplEx with a much larger negative sample set gives a very competitive baseline.

\section{Experiments}
\noindent
{\bf Datasets:} We evaluate on a comprehensive set of five standard KBC datasets - FB15K, WN18, YAGO3-10, FB15K-237 and WN18RR \cite{BordesUGWY2013TransE,yago310-MahdisoltaniBS15,ToutanovaCPPCG2015KgTextDistMult,dettmers2017convolutional}. We retain the exact train, dev and test folds used in previous works. \\

\noindent 
{\bf Metrics: } Link prediction test queries are of the form $(s,r,?)$, which have a gold $o^*$. The cases of $(?,r,o)$ and $(s,r,?)$ are symmetric and receive analogous treatment. KBC models outputs a list of $o\in\mathcal{E}$ ordered (descending) by their scores. We report MRR (Mean Reciprocal Rank) and the fraction of queries where $o^*$ is recalled within rank 1 and rank 10 (HITS). The \emph{filtered} evaluation \cite{GarciaBU2015TransEcompose} removes valid, train or test tuples ranking above $(s,r,o^*)$ to prevent unreasonable model penalization (for predicting another correct answer). \\

\begin{table*}[th!]
\resizebox{\textwidth}{!}{
\centering
\begin{tabular}{c}
\begin{tabular}{|c|c|c|c|c|c|c|c|c|c|}
\hline
                                          & \multicolumn{3}{c|}{\textbf{FB15k}}                             & \multicolumn{3}{c|}{\textbf{WN18}}                & \multicolumn{3}{c|}{\textbf{YAGO3-10}}            \\ \hline
\textbf{Method}                           & \textbf{MRR}               & \textbf{HITS@1} & \textbf{HITS@10} & \textbf{MRR} & \textbf{HITS@1} & \textbf{HITS@10} & \textbf{MRR} & \textbf{HITS@1} & \textbf{HITS@10} \\ \hline
\textbf{SimplE}                        & 0.73                       & 0.65            & 0.86             & 0.95         & 0.94            & 0.95             & ----         & ----            & ----            \\ \hline
\textbf{\sysS{}}                          & 0.85                       & 0.82            &  {\bf 0.91}      & {\bf 0.95}   & {\bf 0.96}      & 0.95             & 0.56         & 0.49            & 0.69            \\ \hline
\textbf{Complex}                          & 0.81                       & 0.75            & {\bf 0.91}       & 0.94         & 0.93            & 0.95             & 0.51         & 0.40            & 0.63          
\\ \hline
\textbf{\sys{}}                           & {\bf 0.86}                 & {\bf 0.83}      &  {\bf 0.91}      & {\bf 0.95}   & 0.95            & {\bf 0.96}       & {\bf 0.58}   & {\bf 0.50}      & {\bf 0.71}   
\\ \hline \hline
\textbf{Complex-N3}                       & {\bf 0.86}                 & {\bf 0.83}      & {\bf 0.91}       & {\bf 0.95}   & 0.94            & {\bf 0.96}       & {\bf 0.58}   & {\bf 0.50}      & {\bf 0.71}   
\\ \hline
\end{tabular}
\\
(a)\\
\begin{tabular}{|c|c|c|c|c|c|c|}
\hline
                                 & \multicolumn{3}{c|}{\textbf{FB15k-237}}                          & \multicolumn{3}{c|}{\textbf{WN18RR}}                             \\ \hline
\textbf{Method}                  & \textbf{MRR}                & \textbf{HITS@1} & \textbf{HITS@10} & \textbf{MRR}                & \textbf{HITS@1} & \textbf{HITS@10} \\ \hline
\textbf{SimplE}               & 0.23                        & 0.15            & 0.40             & 0.42                        & 0.40            & 0.46             \\ \hline
\textbf{\sysS{}}                 & 0.34                        & 0.25            & 0.53             & 0.46                        & 0.43            & 0.52             \\ \hline
\textbf{Complex}                 & 0.31                        & 0.22            & 0.51             & 0.42                        & 0.40            & 0.47             \\ \hline
\textbf{\sys{}}                  & 0.35                        & 0.26            & 0.54             & 0.47                        & {\bf 0.46}      & 0.53             \\ \hline
\hline
\textbf{Complex-N3}              & {\bf 0.37}                  & {\bf 0.27}      & {\bf 0.56}       & {\bf 0.49}                  & 0.44            & {\bf 0.58}       \\ \hline

\end{tabular}\\
(b)\\

\end{tabular}}
\caption{Table (a) and (b) reports performance of popular KBC models along with their full negative sample trained counterparts - \sys{} and \sysS{} on five commonly used benchmark datasets for KBC. All model parameters are of same range. \sys{} shows near SOTA performance on all datasets. Models which use all entities as negative samples during training, are indistinguishable or slightly worse. 
} \label{model_perfromance}
\end{table*}

\noindent
{\bf Implementation details: }
We reused the original implementations and the best hyper-parameters released for RotatE \cite{RotatESunDNT19}.
We re-implemented \complex{} \cite{TrouillonWRGB2016Complex}, CX-N3 \cite{Lacroix2018canonical}, SimplE \cite{kazemi2018simple} in PyTorch. AdaGrad was used for fitting model weights, run for up to 1000 epochs, with early stopping on the dev fold to prevent overfitting. 

\noindent
In our experiments we calibrated all models to have similar number of parameters. \complex{}, CX-N3, \sys{}, \sysS{} use 2000-dimension vectors (1000 dimension on Yago3-10). 

\noindent
All models except CX-N3 use L2 regularization. CX-N3 uses L3 regularization.
\noindent
The ranges of the hyperparameters for the grid search are as follows: regularization coefficient \{1,0.1,0.01,0.001,0.0001,0.00001\}, learning rate \{0.5,0.1,0.01,0.001,0.0001\}, batch size \{100,200,500,1000,2000\}.\\ 
\subsection{Link prediction performance}
This section demonstrates how KBC model performance improves when trained using all possible entities as negative samples.
\subsubsection{Multiplicative models}
Table \ref{model_perfromance} shows multiplicative models when trained with all possible entities as negative samples (\sys{}, \sysS) significantly improve over the same model trained with a small set of negative samples. Note that these models use 1-N scoring for efficient computing the scores. 

Interestingly, models trained with all possible entities as negative samples show near similar performance, providing additional evidence against the value of new variations proposed in form of model architecture, problem reformulation, and regularization. A baseline model trained with all entities as negative sample - \sys{}, shows near SOTA performance, making it a strong baseline.

\subsubsection{Translation models}
\begin{table}[]
\centering
\begin{small}
\begin{tabular}{|c|c|c|c|c|c|}
\hline
          & FB15k & WN18 & YAGO3-10 & FB15k-237 & WN18RR \\ \hline
RotatE    & 0.61      &  0.94     &  0.37        &   0.29        & 0.45        \\ \hline
RotatE-V2 & \bf 0.64      &  \bf0.95     & \bf 0.40         &   \bf 0.32         &  0.45      \\ \hline
\end{tabular}
\caption{Performance (MRR) improvement for RotatE (100-dim) by scoring against all entities while training (instead of negative sampling)}
\label{tab:rotatePerf}
\end{small}
\end{table}

As pointed out in Section \ref{baseline:sec:addvsmul}, 1-N scoring is difficult to scale for models such as RotatE. To demonstrate the usefulness of using all negative samples for training, we train RotatE for a reduced dimension (100). To overcome the memory challenges of training the model on a single 12GB GPU, we train the model by accumulating gradients over multiple batches.  

The results are reported in Table \ref{tab:rotatePerf}. Here, RotatE refers to the model trained with 256 negative samples, whereas RotatE-V2 refers to the model trained with all entities as negative samples. We find that RotatE-V2 shows a significant improvement (upto 3 pt MRR) for FB15k, FB15k-237 and YAGO3-10, whereas for WN18 and WN18RR the model gives a slightly improved or similar performance.

\subsection{Influence of Negative Samples}
\begin{figure}[]
\begin{center}
\includegraphics[width=0.5\textwidth,trim={6.7cm 6.0cm 11.8cm 2.3cm},clip]{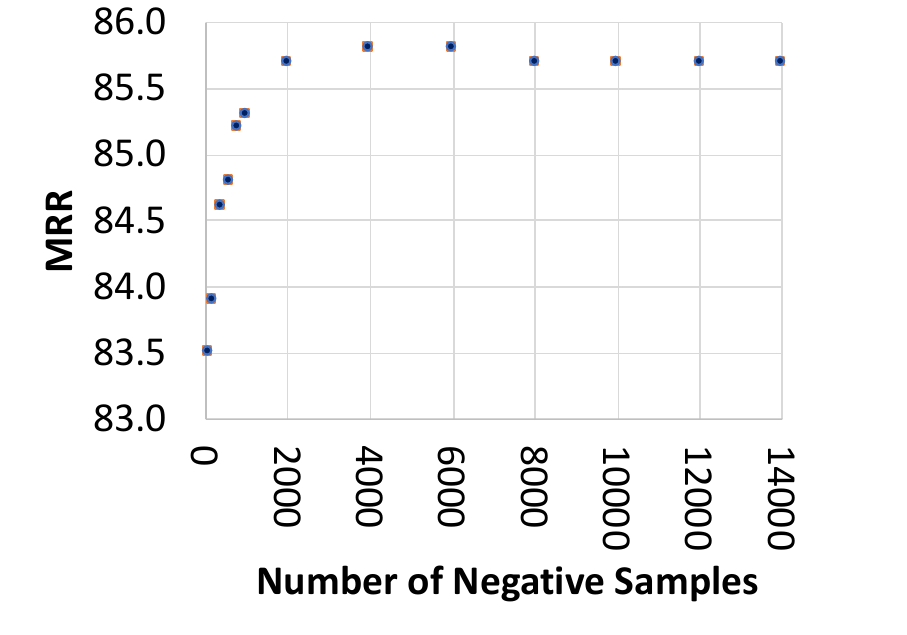}
\end{center}
\caption{Influence on test MRR of number of negative samples used per positive training example.}
\label{Fig:negsample-mrr}
\end{figure}

In this part we investigate model performance with increasing number of negative samples per positive example. We report the performance of Complex model on FB15k dataset for this experiment. We vary the number of negative samples in \{100, 200, 400, 600, 800, 1k, 2k, 4k, 6k, 8k, 10k, 12k,14k\}. 
1-N scoring enables to compute score of all possible entities for query $(e_1, r, *)$ efficiently. We subsample a set of entities (negative examples) for each batch and compute approximate softmax, for this experiment.\\ 
Figure \ref{Fig:negsample-mrr} shows that the performance of Complex sharply improves with increasing number of negative samples (in the beginning) and stabilizes around 2000 negative samples.

\section{Discussion}

The lessons we have learnt from our observations above may be summarized as:
\begin{itemize}
\item
As long as memory footprint is manageable, all KBC models should use large (perhaps even exhaustive) negative samples and vectorized evaluation of contrastive loss --- this most often leads to superior predictive accuracy.

\item
For models where vectorized evaluation of contrastive loss cannot be easily implemented for large negative samples (such as RotatE), gradient accumulation over multiple batches can be used at the cost of increased training time.  This may still lead to better accuracy.

\item
Switching $L_1$ to $L_2$ norms is a tempting possibility to enable fast vectorized evaluation of contrastive loss while keeping memory footprint minimal.  Unfortunately, training TransE and RotatE using $L_2$ norm resulted in visible drop in accuracy.  The MRR score of TransE dropped by 7.8 points and RotatE by 6.5 points on FB15k.  This takes away the possibility of optimizing these models for vectorized contrastive loss evaluation with low memory footprint.

\item
On the positive side, \sys{} --- plain old ComplEx with very large negative samples --- is efficiently trainable and turns out to be an extremely competitive baseline.  In fact, it largely wipes out the benefit from several models proposed since.
\end{itemize}

We are not alone in pointing out the last item above.  \newcite{baseline-KadlecBK17} and \newcite{trick-iclr20-Ruffinelli} undertook an extensive exercise in tuning hyperparameters such as embedding dimensions, learning rate, batch size, regularization penalty, etc., and came to the same conclusion: the apparent accuracy gains from a new model architecture must be carefully assessed against relatively minor-looking but still significant modifications to hyperparameters in well-established models.  Our experiments with negative sample size adds another weapon in the arsenal of old baselines that age well.

\section{Conclusion}
In this paper, we performed an extensive re-examination of recent KBC techniques. We find that models can significantly benefit from using large (exhaustive) negative samples while training. The relative performance gaps between models trained in this manner are small. 
Moreover \sys{} showed SOTA or near SOTA performance on all datasets, making it a strong baseline for other models to use. 

\section*{Acknowledgements}
\label{sec:ack}
This work is supported by IBM AI Horizons Network grant, an IBM SUR award, grants by Google, Bloomberg and 1MG, and a Visvesvaraya faculty award by Govt. of India. We thank IIT Delhi HPC facility for compute resources. Soumen Chakrabarti is supported by grants from IBM and Amazon.

\bibliographystyle{coling}
\bibliography{kbc,voila}

\appendix

\end{document}